\documentclass[10pt,twocolumn,letterpaper]{article}

\usepackage{cvpr}
\usepackage{times}
\usepackage{epsfig}
\usepackage{graphicx}
\usepackage{amsmath}
\usepackage{amssymb}
\usepackage{booktabs}
\usepackage{epstopdf,url}
\usepackage{multirow}
\usepackage{caption}

\usepackage{bm}

\usepackage[breaklinks=true,bookmarks=false]{hyperref}

\cvprfinalcopy 

\def\cvprPaperID{85} 

\ifcvprfinal\fi
\begin{document}

\title{DCAN: Deep Contour-Aware Networks for Accurate Gland Segmentation}

\author{Hao Chen
\and Xiaojuan Qi
\and Lequan Yu
\and Pheng-Ann Heng
\and
Department of Computer Science and Engineering\\
The Chinese University of Hong Kong\\
{\tt\small \{hchen,~xjqi,~lqyu,~pheng\}@cse.cuhk.edu.hk}
}

\maketitle

\begin{abstract}
The morphology of glands has been used routinely by pathologists to assess the malignancy degree of adenocarcinomas.
Accurate segmentation of glands from histology images is a crucial step to obtain reliable morphological statistics for quantitative diagnosis.
In this paper, we proposed an efficient deep contour-aware network (DCAN) to solve this challenging problem under a unified multi-task learning framework.
In the proposed network, multi-level contextual features from the hierarchical architecture are explored with auxiliary supervision for accurate gland segmentation.
When incorporated with multi-task regularization during the training, the discriminative capability of intermediate features can be further improved.
Moreover, our network can not only output accurate probability maps of glands, but also depict clear contours simultaneously for separating clustered objects, which further boosts the gland segmentation performance.
This unified framework can be efficient when applied to large-scale histopathological data without resorting to additional steps to generate contours based on low-level cues for post-separating.
Our method won the 2015 MICCAI Gland Segmentation Challenge out of 13 competitive teams, surpassing all the other methods by a significant margin.
\end{abstract}

\section{Introduction}
\begin{figure}
\centering
  \includegraphics[width=.95\linewidth]{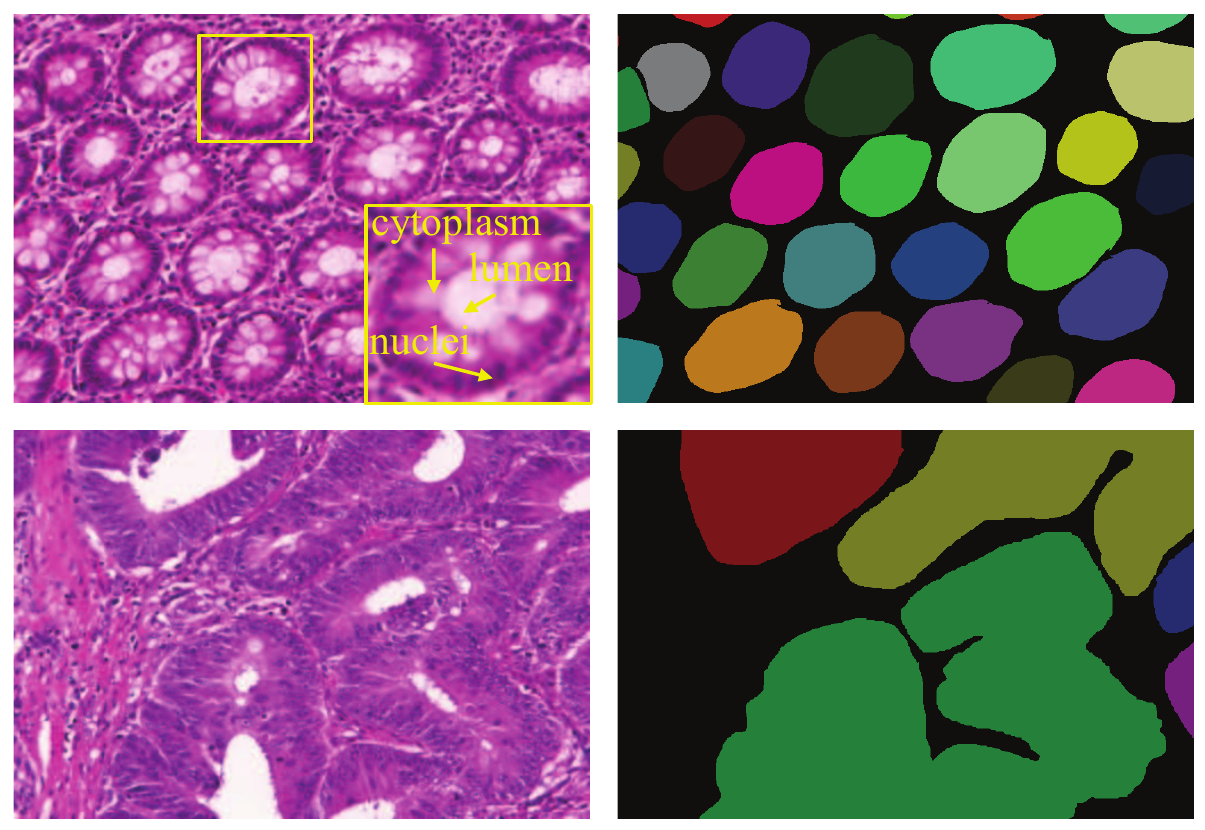}
\caption{Examples of gland segmentation in benign (top row) and malignant (bottom row) cases. From left to right columns show the original images (stained with hematoxylin and eosin) and annotations by pathologists (individual objects are denoted by different colors), respectively.}
\label{fig:example}
\end{figure}

Normally, a typical gland is composed of a lumen area forming the interior tubular structure and epithelial cell nuclei surrounding the cytoplasm, as illustrated in Figure~\ref{fig:example} (top left).
Malignant tumours arising from glandular epithelium, also known as adenocarcinomas, are the most prevalent form of cancer.
In the routine of histopathological examination, the morphology of glands has been widely used for assessing the malignancy degree of several adenocarcinomas, including breast~\cite{elston1991pathological}, prostate~\cite{gleason1992histologic}, and colon~\cite{fleming2012colorectal}. 
Accurate segmentation of glands is one crucial pre-requisite step to obtain reliable morphological statistics that indicate the aggressiveness of tumors. 
Conventionally, this is performed by expert pathologists who evaluate the structure of glands in the biopsy samples.
However, manual annotation suffers from issues such as limited reproducibility, considerable efforts, and time-consuming.
With the advent of whole slide imaging, large-scale histopathological data need to be analyzed.
Therefore, automatic segmentation methods are highly demanded in clinical practice to improve the efficiency as well as reliability and reduce the workload on pathologists.

Nevertheless, this task is quite challenging for several reasons. %
First, there is a huge variation of glandular morphology depending on the different histologic grades as well as from one disease to another.
Figure~\ref{fig:example} (left column) shows the large difference of glandular structures between benign and malignant cases from colon tissues.
Second, the existence of touching glands in tissue samples makes it quite hard for automated methods to separate objects individually.
Third, in the malignant cases such as moderately and poorly differentiated adenocarcinomas, the glandular structures are seriously degenerated, as shown in Figure~\ref{fig:example} (bottom left).
Therefore, methods utilizing the prior knowledge with glandular regularity are prone to fail in such cases~\cite{sirinukunwattana2015stochastic}.
In addition, the variation of tissue preparation procedures such as sectioning and staining can cause deformation, artifacts and inconsistency of tissue appearance, which can impede the segmentation process as well.

In the last few years, many researchers have devoted their efforts to addressing this challenging problem and achieved a considerable progress.
However, obvious performance gap is still observed between the results given by the algorithms and annotations from pathologists.
Broadly speaking, previous studies in the literature can be categorized into two classes:
(1) pixel based methods.
For this kind of method, various hand-crafted features including texture, color, morphological cues and Haar-like features were utilized to detect the glandular structure from histology images~\cite{diamond2004use,wu2005segmentation,doyle2006boosting,sirinukunwattana2015novel,tabesh2007multifeature,nguyen2012structure,jacobs2014gleason,sabata2010automated};
(2) structure based methods. 
Most of approaches in this category take advantage of prior knowledge about the glandular structure, such as graph based methods~\cite{altunbay2010color,gunduz2010automatic}, glandular boundary delineation with geodesic distance transform~\cite{fakhrzadeh2012analyzing}, polar space random field model~\cite{fu2014novel}, stochastic polygons model~\cite{sirinukunwattana2015stochastic}, etc. 
Although these methods achieved promising results in cases of adenoma and well differentiated (low grade) adenocarcinoma, they may fail to achieve satisfying performance in malignant subjects, where the glandular structures are seriously deformed.
Recently, deep neural networks are driving advances in image recognition related tasks in computer vision~\cite{he2015deep,dai2015boxsup,chen2014semantic,long2015fully,qi2015semantic,bertasius2015high} and medical image computing~\cite{dhungel2015deep,ronneberger2015u,roth2015deeporgan,chen2015spine,dou2016automatic}.
The most relevant study to our work is the~\emph{U-net} that designed a U-shaped deep convolutional network for biomedical image segmentation and won several grand challenges recently~\cite{ronneberger2015u}.

In this paper, we propose a novel deep contour-aware network to solve this challenging problem.
Our method tackles three critical issues for gland segmentation.
First, our method harnesses multi-level contextual feature representations in an end-to-end way for effective gland segmentation. 
Leveraging the fully convolutional networks, it can take an image as input and output the probability map directly with one single forward propagation. Hence, it's very efficient when applied to large-scale histopathological image analysis.
Second, because our method doesn't make an assumption about glandular structure, it can be easily generalized to biopsy samples with different histopathological grades including benign and malignant cases.
Furthermore, instead of treating the segmentation task independently, our method investigates the complementary information, \ie, gland objects and contours, under a multi-task learning framework.
Therefore, it can simultaneously segment the gland and separate the clustered objects into individual ones, especially in benign cases with existence of touching glands.
Extensive experimental results on the benchmark dataset of~\emph{2015 MICCAI Gland Segmentation Challenge} corroborated the effectiveness of our method, yielding much better performance than other advanced methods.


\begin{figure}
\centering
  \includegraphics[width=1.0\linewidth]{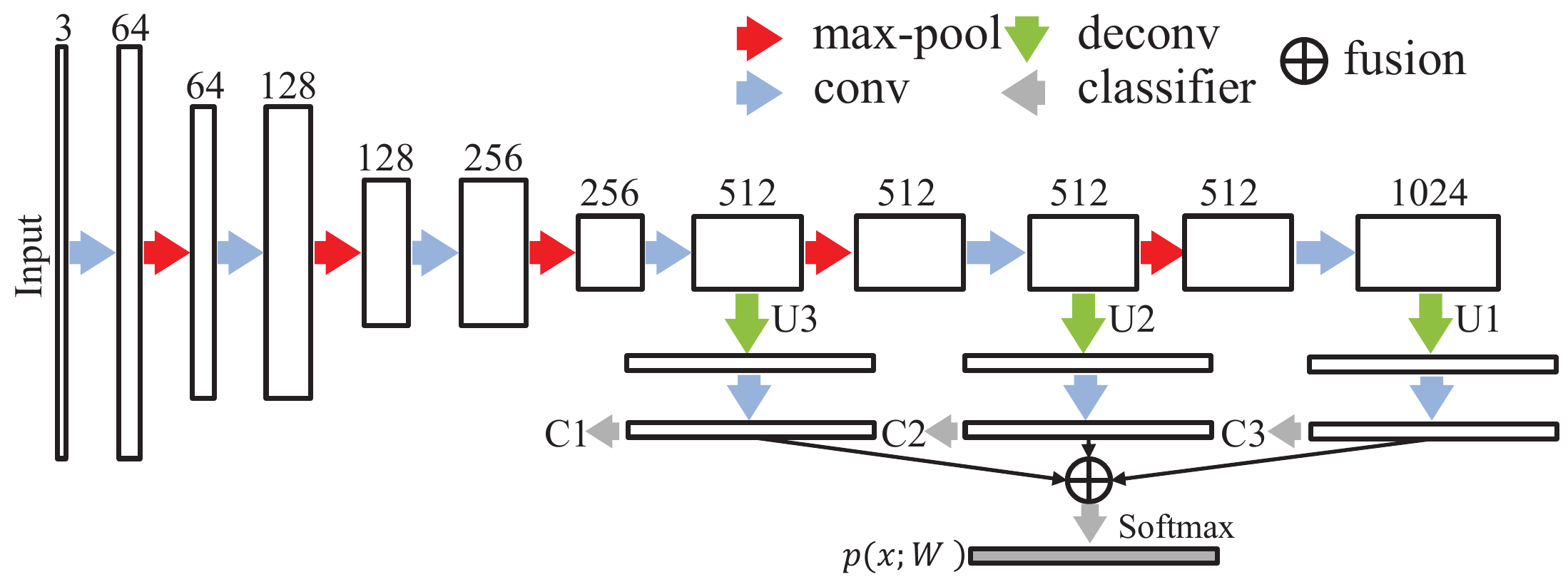}
\caption{The schematic illustration of FCN with multi-level contextual features.}
\label{fig:fcn}
\end{figure}

\begin{figure*}
\centering
  \includegraphics[width=.81\linewidth]{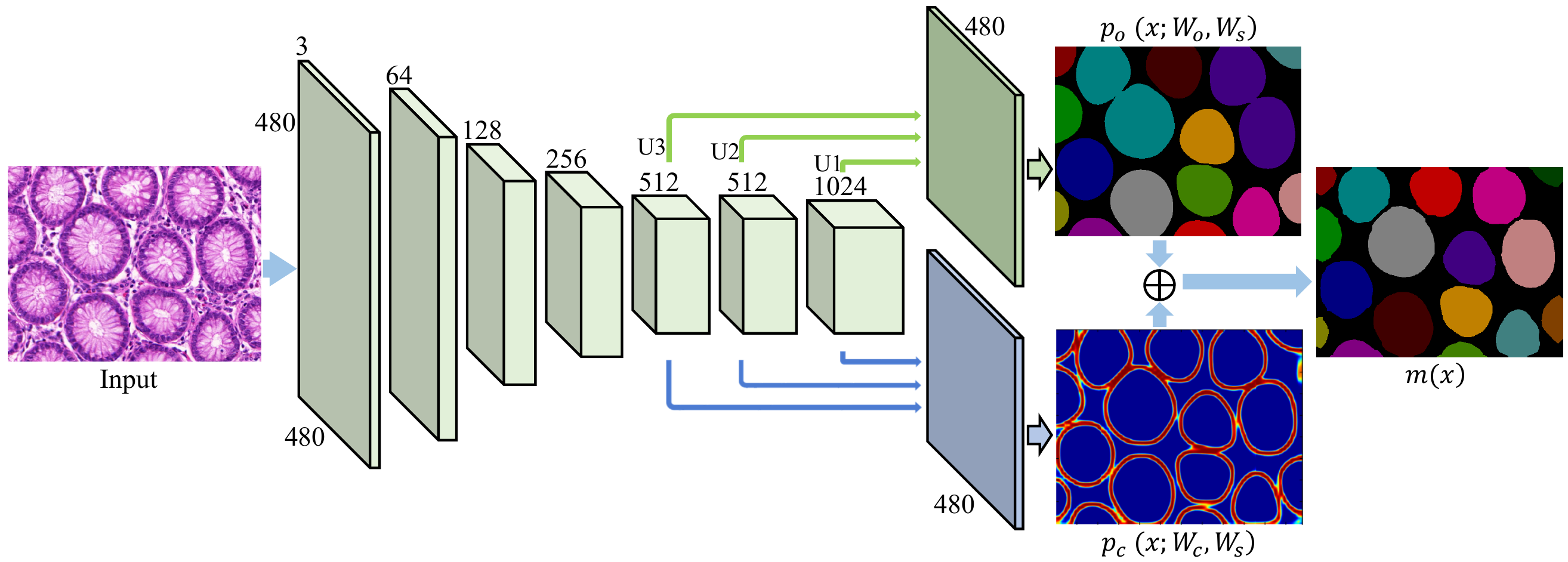}
\caption{The overview of the proposed deep contour-aware network.}
\label{fig:framework}
\end{figure*}

\section{Method}
In this section, we describe in detail the formulation of our proposed deep contour-aware network for accurate gland segmentation.
We start by introducing the fully convolutional network (FCN) for end-to-end training.
Furthermore, we propose to harness the multi-level contextual features with auxiliary supervision for generating good likelihood maps of glands.
Then we elaborate the deep contour-aware network drawn from FCN for effective gland segmentation by fusing the complementary information of objects and contours.
In order to mitigate the challenge of insufficient training data, we employ the transfer learning approach by exploiting the knowledge learned from cross domains to further improve the performance.

\subsection{FCN with multi-level contextual features}
\label{subsec:fcn}
Fully convolutional networks achieved the state-of-the-art performance on image segmentation related tasks~\cite{chen2014semantic,long2015fully}. 
Such great success is mostly attributed to the outstanding capability in feature representation for dense classification.
The whole network can be trained in an end-to-end (image-to-image) way, which takes an image as input and outputs the probability map directly. 
The architecture basically contains two modules including downsampling path and upsampling path.
The downsampling path contains convolutional and max-pooling layers, which are extensively used in the convolutional neural networks for image classification tasks~\cite{ciresan2012deep,krizhevsky2012imagenet}.
The upsampling path contains convolutional and deconvolutional layers (backwards strided convolution~\cite{long2015fully}), which upsample the feature maps and output the score masks. The motivation behind this is that the downsampling path aims at extracting the high level abstraction information, while the upsampling path predicting the score masks in a pixel-wise way.

The classification scores from FCN are established based on the intensity information from the given receptive field.
However, the network with single receptive field size cannot handle the large variation of gland shape properly.
For example, as shown in Figure~\ref{fig:example}, a small receptive field (\eg, $150\times 150$) is suitable for normal glands in benign cases, while malignant cases usually need a large receptive field since the gland shape in adenocarcinomas are degenerated and elongated, hence enclosing larger contextual information can help to eliminate ambiguity, suppress the interior tubular structure, and improve the recognition performance.
Therefore, based on the FCN, we push it further by harnessing multi-level contextual feature representations, which include different levels of contextual information, \ie, intensities appearing in various sizes of receptive field.
The schematic illustration of FCN with multi-level contextual feature representations can be seen in Figure~\ref{fig:fcn}.
Specifically, the architecture of neural network contains a number of convolutional layers, 5 max-pooling layers for downsampling and 3 deconvolutional layers for upsampling.
With the network going deeper, the size of receptive field is becoming larger. 
Derived from this, the upsampling layers are designed deliberately by considering the requirement of different receptive field sizes.
They upsample the feature maps and make predictions based on the contextual cues from given receptive field.
Then these predictions are fused together with a summing operation and final segmentation results based on multi-level contextual features are generated after softmax classification.

Direct training a network with such a large depth may fall into a local minima.
Inspired by previous studies on training neural networks with deep supervision~\cite{leedeeply,xie2015holistically,hao2016deep},
weighted auxiliary classifiers C1-C3 are added into the network to further strengthen the training process, as shown in Figure~\ref{fig:fcn}.
This can alleviate the problem of vanishing gradients with auxiliary supervision for encouraging the back-propagation of gradient flow.
Finally, the FCN with multi-level contextual features extracted from input $I$ can be trained by minimizing the overall loss $\mathcal{L}$, \ie, a combination of auxiliary loss $\mathcal{L}_a (I;W)$ with corresponding discount weights $w_a$ and data error loss $\mathcal{L}_e (I;W)$ between the predicted results and ground truth annotation, as shown following:
\begin{gather}\label{fcn_loss}
  \mathcal{L}(I;W) = \lambda \psi(W) + \sum_a w_a \mathcal{L}_a(I;W)+ \mathcal{L}_e(I;W)
\end{gather}
where $W$ denotes the parameters of neural network and $\psi(W)$ is the regularization term with hyperparameter $\lambda$ for balancing the tradeoff with other terms.

\subsection{Deep contour-aware network}
By harnessing the multi-level contextual features with auxiliary supervision, the network can produce good probability maps of gland objects.
However, it's still quite hard to separate the touching glands by leveraging only on the likelihood of gland objects due to the essential ambiguity in touching regions.
This is rooted in the downsampling path causing spatial information loss along with feature abstraction.
The boundary information formed by epithelial cell nuclei provides good complementary cues for splitting objects.
To this end, we propose a deep contour-aware network to segment the glands and separate clustered objects into individual ones.

The overview of the proposed deep contour-aware network can be seen in Figure~\ref{fig:framework}.
Instead of treating the gland segmentation task as a single and independent problem, we formulate it as a multi-task learning framework by exploring the complementary information, which can infer the results of gland objects and contours simultaneously.
Specifically, the feature maps are upsampled with two different branches (green and blue arrows shown in the figure) in order to output the segmentation masks of gland objects and contours, respectively.
In each branch, the mask is predicted by FCN with multi-level contextual features as illustrated in Section~\ref{subsec:fcn}.
During the training process, the parameters of downsampling path $W_s$ are shared and updated for these two tasks jointly, while the parameters of upsampling layers for two individual branches (denoted as $W_o$ and $W_c$) are updated independently for inferring the probability of gland objects and contours, respectively.
Therefore, the feature representations through the hierarchical structure can encode the information of segmented objects and contours at the meantime.
Note that the network with multiple tasks is optimized together in an end-to-end way.
This joint multi-task learning process has several advantages.
First, it can increase the discriminative capability of intermediate feature representations with multiple regularizations on disentangling subtly correlated tasks~\cite{zhang2014facial}, hence improve the robustness of segmentation performance. 
Second, in the application of gland segmentation, the multi-task learning framework can also provide the complementary contour information that serves well to separate the clustered objects.
This can improve the object-level segmentation performance significantly, especially in benign histology images where touching gland objects often exist.
When dealing with large-scale histopathological data, this unified framework can be quite efficient. With one forward propagation, it can generate the results of gland objects and contours simultaneously instead of resorting to additional post-separating steps by generating contours based on low-level cues~\cite{gunduz2010automatic,wu2005segmentation}.

In the training process, the discount weights $w_a$ from auxiliary classifiers are decreased until marginal values with the number of iterations increasing, therefore we dropped these terms in the final loss for simplicity.
Finally the training of network is formulated as a per-pixel classification problem regarding the ground truth segmentation masks including gland objects and contours, as shown following:
\begin{gather}
  \mathcal{L}_{\text{total}}(x;\theta) = {\lambda} \psi(\theta)-
   \sum_{x\in \mathcal{X}} \log p_o(x,\ell_o(x);W_o,W_s)\nonumber \\
    - \sum_{x\in \mathcal{X}} \log p_c(x,\ell_c(x);W_c,W_s)\label{loss}
\end{gather}
where the first part is the $L_2$ regularization term and latter two are the data error loss terms.
$x$ is the pixel position in image space $\mathcal{X}$, $p_o(x,\ell_o(x);W_o,W_s)$ denotes the predicted probability for true label $\ell_o(x)$ (\ie, the index of 1 in one hot vector) of gland objects after softmax classification, and similarly $p_c(x,\ell_c(x);W_c,W_s)$ is the predicted probability for true label $\ell_c(x)$ of gland contours.
The parameters $\theta=\{W_s,W_o,W_c\}$ of network are optimized by minimizing the total loss function $\mathcal{L}_{\text{total}}$ with standard back-propagation.

With the predicted probability maps of gland object $p_o(x;W_c,W_s)$ and contour $p_c(x;W_c,W_s)$ from the deep contour-aware network, these complementary information are fused together to generate the final segmentation masks $m(x)$, defined as:
\begin{gather}\label{fusion}
\resizebox{\linewidth}{!}{ $m(x)=
\begin{cases}
1& ~\text{if}~p_o(x;W_o,W_s)\geq t_o~\text{and}~p_c(x;W_c,W_s)< t_c\\
0 & \text{otherwise}
\end{cases}$}
\end{gather}
where $t_o$ and $t_c$ are the thresholds (set as 0.5 in our experiments empirically).
Then, post-processing steps including smoothing with a disk filter (radius 3), filling holes and removing small areas are performed on the fused segmentation results.
Finally, each connected component is labeled with a unique value for representing one segmented gland.

\subsection{Transfer learning with rich feature hierarchies }
There is a scarcity of medical training data along with accurate annotations in most situations due to the expensive cost and complicated acquisition procedures. 
Compared with the limited data in medical domain, much more training data can be obtained in the field of computer vision.
Previous studies have evidenced that transfer learning in deep convolutional networks can alleviate the problem of insufficient training data~\cite{chen2015automatic,shin2016deep}.
The learned parameters (convolutional filters) in the lower layers of network are general while those in higher layers are more specific to different tasks~\cite{yosinski2014transferable}.
Thus, transfer the rich feature hierarchies with embedded knowledge learned from plausibly related datasets could help to reduce overfitting on limited medical dataset and further boost the performance.

Therefore, we utilized an off-the-shelf model from DeepLab~\cite{chen2014semantic}, which was trained on the PASCAL VOC 2012 dataset~\cite{everingham2010pascal}.
Compared to the small scale dataset (a few hundred images) in gland segmentation, the PASCAL VOC dataset contains more than ten thousand images with pixel-level annotations. 
Leveraging the effective generalization ability of transfer learning in deep neural networks, we initialized the layers in downsampling path with pre-trained weights from the DeepLab model while the rest layers randomly with Gaussian distribution.
Then we fine tuned the whole network on our medical task in an end-to-end way with stochastic gradient descent.
In our experiments, we observed the training process converged much faster (about four hours)  by virtue of the prior knowledge learned from rich dataset than random initialization setting.

\section{Experiments and results}
\subsection{Dataset and pre-processing}
We evaluated our method on the public benchmark dataset of~\emph{Gland Segmentation Challenge Contest} in MICCAI 2015 (also named as Warwick-QU dataset)~\cite{sirinukunwattana2016gland}.
The images were acquired by a Zeiss MIRAX MIDI slide scanner from colorectal cancer tissues with a resolution of $0.62 \mu{m}$/pixel.
They consist of a wide range of histologic grades from benign to malignant subjects.
It's worth noting that poorly-differentiated cases are included to evaluate the performance of algorithms.
The training dataset is composed of 85 images (benign/malignant=37/48) with ground truth annotations provided by expert pathologists.
The testing data contains two sections: Part A  (60 images) for offline evaluation and Part B (20 images) for on-site evaluation.
For the on-site contest, participants must submit their results to the organizers within an hour after data release.
The ground truths of testing data are held out by the challenge organizers for independent evaluation.
The final ranking is based on the evaluation results from testing data Part A and Part B with an equal weight\footnote{Please refer to the challenge website for more details: \url{http://www2.warwick.ac.uk/fac/sci/dcs/research/combi/research/bic/glascontest/}}.
To increase the robustness and reduce overfitting, we utilized the strategy of data augmentation to enlarge the training dataset.
The augmentation transformations include translation, rotation, and elastic distortion (\eg, pincushion and barrel distortions).

\subsection{Implementation details}
Our framework was implemented under the open-source deep learning library Caffe~\cite{jia2014caffe}.
The network randomly crops a $480\times 480$ region from the original image as input and output the prediction masks of gland objects and contours.
The score masks of whole testing image are produced with an overlap-tile strategy.
For the label of contours, we extracted the boundaries of connected components based on the gland annotations from pathologists, then dilated them with a disk filter (radius 3).
In the training phase, the learning rate was set as 0.001 initially and decreased by a factor of 10 when the loss stopped decreasing till $10^{-7}$.
The discount weight $w_a$ was set as 1 initially and decreased  by a factor of 10 every ten thousand iterations until a marginal value $10^{-3}$.
In addition, dropout layers~\cite{hinton2012improving} (dropout rate 0.5) were incorporated in the convolutional layers with kernel size $1\times1$ for preventing the co-adaption of intermediate features.

\begin{figure*}
\centering
  \includegraphics[width=.95\linewidth]{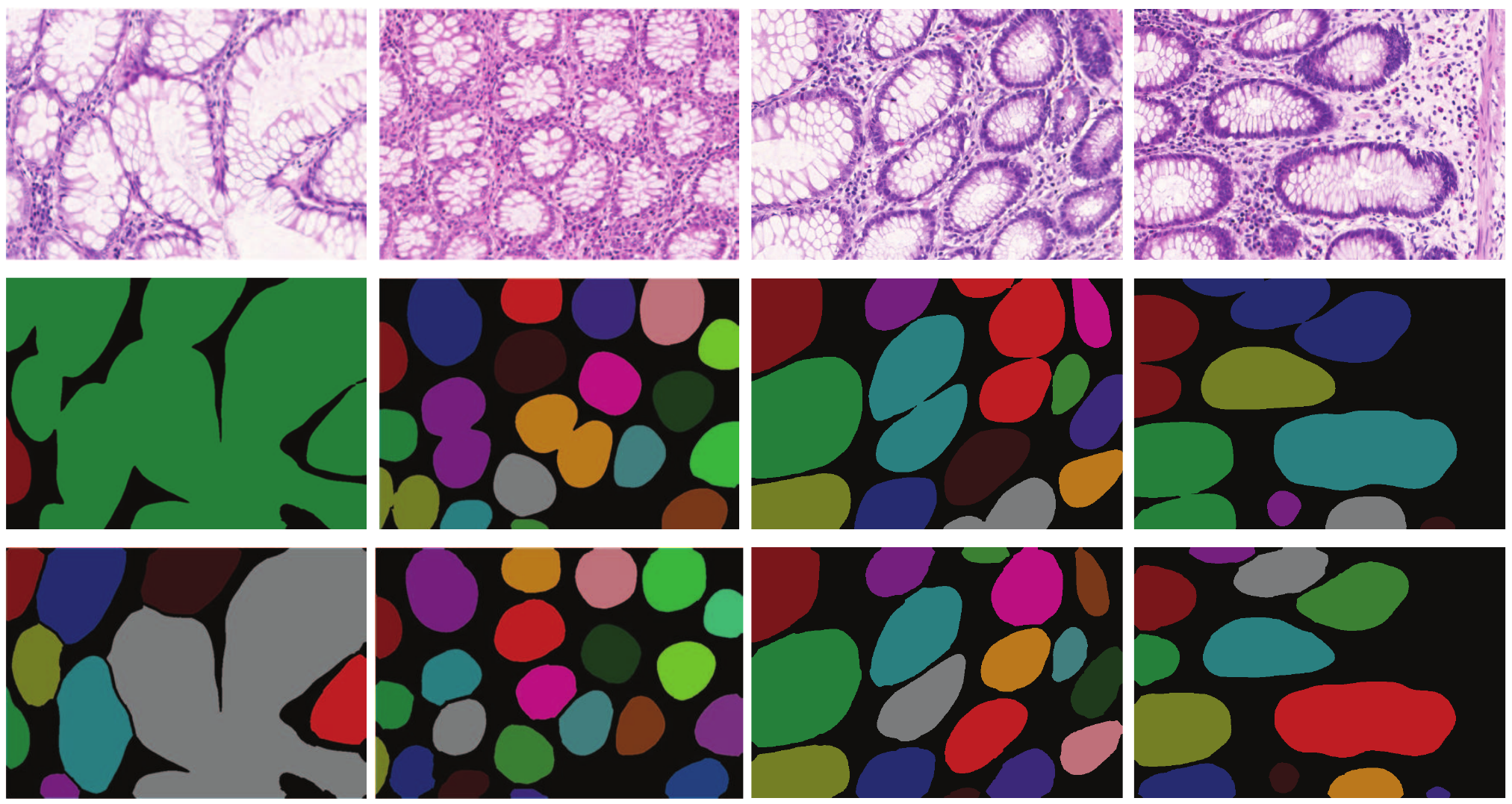}
\caption{Segmentation results of benign cases  (from top to bottom): original images, segmentation results without contour-aware, and segmentation results with contour-aware (different colors denote individual gland objects).}
\label{fig:result1}
\end{figure*}

\begin{figure*}
\centering
  \includegraphics[width=.95\linewidth]{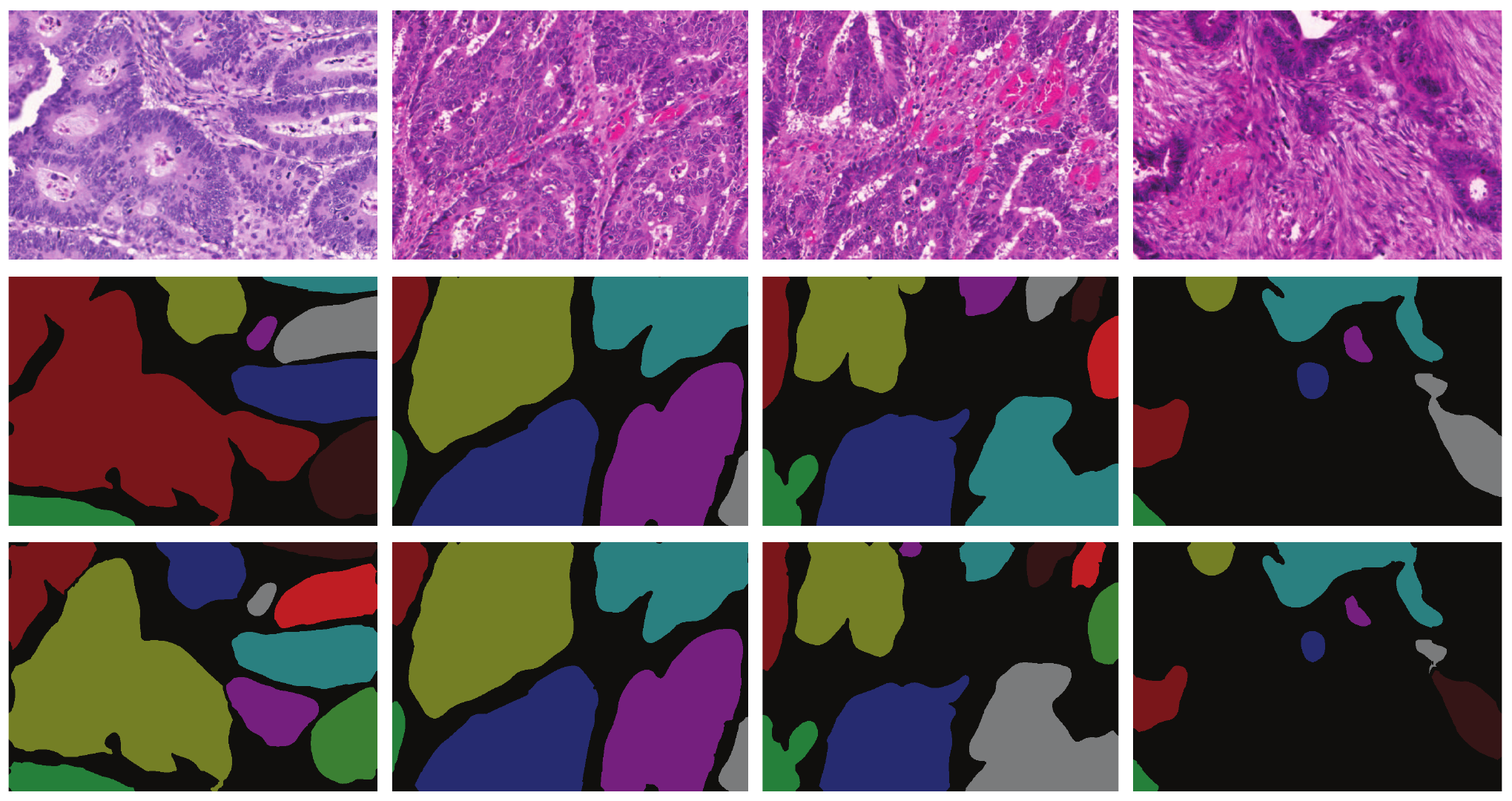} 
\caption{Segmentation results of malignant cases (from top to bottom): original images, segmentation results without contour-aware, and segmentation results with contour-aware  (different colors denote individual gland objects).}
\label{fig:result2}
\end{figure*}

\subsection{Qualitative evaluation}
In order to illustrate the efficacy of our method qualitatively, some segmentation results of testing data are shown in Figure~\ref{fig:result1} (benign cases) and Figure~\ref{fig:result2} (malignant cases), respectively.
For diagnosing the role of complementary contour information (\ie, contour-aware component), we also performed an ablation study and compared the performance of network relying only on the prediction of gland objects.
Qualitative results are shown in Figure~\ref{fig:result1} and Figure~\ref{fig:result2} (middle row).
From the segmentation results we can see that the method leveraging the multi-level contextual features without contour-aware can accurately segment the gland objects in both benign and malignant cases.
However, there are some touching gland objects that cannot be separated. 
The situation is deteriorated when the touching objects are clustered together, as the case shown in the first column of Figure~\ref{fig:result1}.
In comparison, the deep contour-aware network is capable of separating these touching gland objects clearly.
This highlights the superiority of deep contour-aware network by exploring the complementary information under a unified multi-task learning framework qualitatively.

\subsection{Quantitative evaluation and comparison}
The evaluation criteria in the grand challenge includes F1 score, object-level Dice index and Hausdorff distance, which consider the performance of gland detection, segmentation and shape similarity, respectively.
Due to limited submissions in this challenge, we only submitted two entries to probe the performance of our method quantitatively.
They were generated from the deep contour-aware network illustrated in Figure~\ref{fig:framework} without and with fusing the contour-aware results, denoted as \emph{CUMedVision1} and~\emph{CUMedVision2}, respectively.
\\
\textbf{Detection}~~For the gland detection evaluation, the metric F1 score is utilized, which is the harmonic mean of precision $P$ and recall $R$, defined as:
\begin{gather}\label{detection}
  F1 = \frac{2 PR}{P+R},~P=\frac{N_{tp}}{N_{tp}+N_{fp}},~R=\frac{N_{tp}}{N_{tp}+N_{fn}}
\end{gather}
where $N_{tp}$, $N_{fp}$, and $N_{fn} $ denote the number of true positives, false positives, and false negatives, respectively.
According to the challenge evaluation, the ground truth for each segmented object is the object in the manual annotation that has maximum overlap with that segmented object.
A segmented gland object that intersects with at least 50\% of its ground truth is considered as a true positive, otherwise it's considered as a false positive.
A ground truth gland object that has no corresponding segmented object or has less than 50\% of its area overlapped by its corresponding segmented object is considered as a false negative.

The detection results of different methods in this challenge are shown in Table~\ref{table:result_f1}.
Note that all the top 5 entries utilized methods based on the deep convolutional neural networks.
Specially, the method from~\emph{Freiburg} designed a very deep U-shaped network and achieved the best results in several grand challenges~\cite{ronneberger2015u}.
This method also explored the multi-level feature representations by concatenating feature maps from hierarchical layers and weighted loss was utilized to separate the touching objects.

Our submitted entry~\emph{CUMedVision1} without fusing the contour-aware results surpassed all the other methods by a significant margin on testing data Part B, highlighting the strength of FCN with multi-level contextual feature representations for image segmentation.
Our second submitted entry~\emph{CUMedVision2} with contour-aware component achieved the best results on testing data Part A and competitive performance on Part B, which demonstrated the effectiveness of deep contour-aware network on this challenging problem.
From Table~\ref{table:result_f1}, we noticed that all methods yielded relatively lower performance on testing data Part B than Part A.
This mainly comes from the different data distributions.
We observed that benign cases make up about $55\%$ in Part A while most of Part B are malignant cases.
\emph{CUMedVision2} achieved inferior performance (but still competitive compared to other methods) than \emph{CUMedVision1} on Part B.
This arises from the fact that irregular structures in malignant cases can make the gland segmentation more challenging.
For example, the low-contrast between interior tubular structure and stroma as a result of tissue degeneration may make methods relying on epithelial boundary cues more likely fail in such cases.
Nevertheless, our deep contour-aware network ranked first regarding the detection results on all testing data.
\begin{table}
\begin{center}
\begin{tabular}{ | l | l | l | }
\hline
	Method & Part A & Part B \\ \hline
	\bf{CUMedVision2} & \bf{0.9116} & 0.7158 \\
	ExB3 & 0.8958 & 0.7191 \\
	\bf{CUMedVision1} & 0.8680 & \bf{0.7692} \\
	ExB1 & 0.8912 & 0.7027 \\
	ExB2 & 0.8924 & 0.6857 \\
	Freiburg2~\cite{ronneberger2015u} & 0.8702 & 0.6952\\
	CVIP Dundee & 0.8633 & 0.6328 \\
	Freiburg1~\cite{ronneberger2015u} & 0.8340 & 0.6047 \\
	CVML & 0.6521 & 0.5408 \\
	Ching-Wei Wang1& 0.5431 & 0.4790 \\ \hline
\end{tabular}
\caption{The detection results of different methods in 2015 MICCAI Gland Segmentation Challenge (only top 10 entries are shown here and the ranking from top to bottom is made according to the standard competition ranking~\cite{standardrank}).}
\label{table:result_f1}
\end{center}
\end{table}
\\
\textbf{Segmentation}~~Given a set of pixels $G$ annotated as a ground truth object and a set of pixels $S$ segmented as a gland object, Dice index is often employed for segmentation evaluation ${D}(G,S) = 2(|G\cap{S}|)/({|G|+|S|})$.
However, this is not suitable for segmentation evaluation on individual objects.
Instead, an object-level Dice index is utilized and defined as:
\begin{gather}\label{dice}
  {D}_\text{object}(G,S)=\frac{1}{2}\left[\sum_{i=1}^{n_S}\omega_i{D}(G_i,S_i)+\sum_{j=1}^{n_G}\tilde{\omega}_j{D}(\tilde{G}_j,\tilde{S}_j)\right]
\end{gather}
where $S_i$ denotes the $i$th segmented object,
$G_i$ denotes a ground truth object that maximally overlaps $S_i$,
$\tilde{G}_j$ denotes the $j$th ground truth object,
$\tilde{S}_j$ denotes a segmented object that maximally overlaps $\tilde{G}_j $,
$\omega_i=|S_i|/\sum_{m=1}^{n_S}|S_m|, \tilde{\omega}_j=|\tilde{G}_j|/\sum_{n=1}^{n_G}|\tilde{G}_n|$,
$n_S$ and $n_G$ are the total number of segmented objects and ground truth objects, respectively.

The segmentation results of different methods are shown in Table~\ref{table:result_dice}.
We can see that our results~\emph{CUMedVision2} achieved the best performance on testing data Part A and~\emph{CUMedVision1} outperformed all the other advanced methods on Part B.
Similarly, there is around 3\% improvement in Part A and 2\% decrement on Part B in terms of object-level Dice index comparing our method with and without fusing contour-aware results.
By examining some malignant cases, we observed that some inaccurate contours in interior structures may cause the deformed glands fragmented.
One failure example is shown in Figure~\ref{fig:result2} (fourth column), which indicates that contours may over-split the object in some seriously degenerated cases.
In summary, the deep contour-aware network achieved the best segmentation results regarding the object-level Dice index on all testing data, which evidenced the efficacy of our method consistently.
\begin{table}
\begin{center}
\begin{tabular}{ | l | l | l | }
\hline
	Method & Part A & Part B \\ \hline
	\bf{CUMedVision2} & \bf{0.8974} & 0.7810 \\
	ExB1 & 0.8823 & 0.7860 \\
	ExB3 & 0.8860 & 0.7647 \\
	Freiburg2~\cite{ronneberger2015u} & 0.8756 & 0.7856 \\
	\bf{CUMedVision1} & 0.8666 & \bf{0.8001} \\
	ExB2 & 0.8844 & 0.7542 \\
	Freiburg1~\cite{ronneberger2015u} & 0.8745 & 0.7832 \\
	CVIP Dundee & 0.8698 & 0.7152 \\
	LIB & 0.8012  & 0.6166 \\
	CVML & 0.6444 & 0.6543 \\ \hline
\end{tabular}
\caption{The segmentation results of different methods in 2015 MICCAI Gland Segmentation Challenge.}
\label{table:result_dice}
\end{center}
\end{table}
\textbf{Shape similarity}~~
The shape similarity is measured by using the Hausdorff distance between the shape of segmented object and that of the ground truth object, defined as:
\begin{gather}\label{hausdorff}
  H(G,S) = \max\{\sup_{x\in{G}} \inf_{y\in{S}}\|x-y\|,\sup_{y\in{S}}\inf_{x\in{G}}\|x-y\|\}
\end{gather}
Likewise, an object-level Hausdorff is employed: 
\begin{gather}\label{dice}
  {H}_\text{object}(G,S)=\frac{1}{2}\left[\sum_{i=1}^{n_S}\omega_i{H}(G_i,S_i)+\sum_{j=1}^{n_G}\tilde{\omega}_j{H}(\tilde{G}_j,\tilde{S}_j)\right]
\end{gather}

The shape similarity results of different methods are shown in Table~\ref{table:result_hausdorff}.
Our results~\emph{CUMedVision2} from deep contour aware network achieved the smallest Hausdorff distance (the only one less than 50 pixels), outperforming other methods by a significant margin on testing data Part A.
In addition, the results of~\emph{CUMedVision1} is comparable to the best results from~\emph{ExB1} regarding the shape similarity on Part B.
\begin{table}
\begin{center}
\begin{tabular}{ | l | l | l | }
\hline
	Method & Part A & Part B \\ \hline
	Freiburg2~\cite{ronneberger2015u} & 57.0932 & 148.4630 \\
	Freiburg1~\cite{ronneberger2015u} & 57.1938 & 146.6065 \\
	\bf{CUMedVision2} & \bf{45.4182} & 160.3469 \\
	\bf{ExB1} & 57.4126 & \bf{145.5748} \\
	ExB2 & 54.7853 & 187.4420 \\
	ExB3 & 57.3500 & 159.8730 \\
	CUMedVision1 & 74.5955 & 153.6457 \\
	CVIP Dundee & 58.3386 & 209.0483 \\
	LIB & 101.1668  & 190.4467 \\
	CVML & 155.4326 & 176.2439 \\ \hline
\end{tabular}
\caption{The shape similarity results of different methods in 2015 MICCAI Gland Segmentation Challenge.}
\label{table:result_hausdorff}
\end{center}
\end{table}
\\
\textbf{Overall results}~~
For the overall results, each team is assigned three ranking numbers for each part of testing data based on the three criteria mentioned above, one ranking number per criterion, using a standard competition ranking~\cite{standardrank}.
The sum score of these numbers is used for the final ranking, \ie, a smaller score stands for better overall segmentation results.
The final ranking can be seen in Table~\ref{table:result_final} (only top 10 entries are shown).
Although there is a side-effect with contour-aware component in some malignant cases, our deep contour-aware network yielded the best performance in terms of overall results out of 13 teams, outperforming all the other advanced methods by a significant margin.
One straightforward way to refrain from the side-effect is to classify the histopathological images into benign and malignant cases first, then segment the image with contour-aware component or not depending on the classification results.
This may enlighten other researchers for more advanced fusion algorithms.
\begin{table*}
\centering
\begin{tabular}{ | l | c | c | c | c | c | c | c | c | }
 \hline
 \multirow{2}{*}{Method} &
 \multicolumn{6}{c|}{Ranking score} &
 \multirow{2}{*}{Sum score}&
 \multirow{2}{*}{Final ranking} \\
 \cline{2-7}
 &F1 A & F1 B & Dice A & Dice B &Hausdorff A & Hausdorff B & &\\
 \hline
	\bf{CUMedVision2}  & \bf{1} & 3 & \bf{1} & 5 & \bf{1} & 6 & \bf{17} & \bf{1} \\ \hline
	\bf{ExB1} & 4 & 4 & 4 & 2 & 6 & \bf{1} & 21 & 2 \\ \hline
	ExB3 & 2 & 2 & 2 & 6 & 5 & 5 & 22 & 3 \\ \hline
	Freiburg2~\cite{ronneberger2015u} & 5 & 5 & 5 & 3 & 3 & 3 & 24 & 4 \\ \hline
	\bf{CUMedVision1} & 6 & \bf{1} & 8 & \bf{1} & 8 & 4 & 28 & 5 \\ \hline
	ExB2 & 3 & 6 & 3 & 7 & 2 & 8 & 29 & 6 \\ \hline
	Freiburg1~\cite{ronneberger2015u} & 8 & 8 & 6 & 4 & 4 & 2 & 32 & 7 \\ \hline
	CVIP & 7 & 7 & 7 & 8 & 7 & 10 & 46 & 8 \\ \hline
	CVML & 10 & 9 & 11 & 9 & 11 & 7 & 57 & 9 \\ \hline
	LIB & 9 & 16 & 9 & 12 & 9 & 9 & 64 & 10 \\ \hline
\end{tabular}
\caption{The final ranking of different methods in 2015 MICCAI Gland Segmentation Challenge (A and B denote the part of testing data, only top 10 entries are shown here).}
\label{table:result_final}
\end{table*}
\subsection{Computation cost}
It took about four hours to train the deep contour-aware network on a workstation with 2.50 GHz Intel(R) Xeon(R) E5-1620 CPU and a NVIDIA GeForce GTX Titan X GPU.
Leveraging the efficient inference of fully convolutional architecture, the average time for processing one testing image with size $755\times522$ was about 1.5 seconds, which was much faster than other methods~\cite{sirinukunwattana2015stochastic,gunduz2010automatic} in the literature. 
Considering large-scale histology images are demanded for prompt analysis with the advent of whole slide imaging, the fast speed implies the possibility of our method in clinical practice.
\section{Conclusions}
In this paper, we have presented a deep contour-aware network that integrates multi-level contextual features to accurately segment glands from histology images.
Instead of learning gland segmentation in isolation, we formulated it as a unified multi-task learning process by harnessing the complementary information, which helps to further separate the clustered gland objects efficiently.
Extensive experimental results on the benchmark dataset with rich comparison results demonstrated the outstanding performance of our method.
In the future work, we will optimize the method and investigate its capability on large-scale histopathological dataset.

\section*{Acknowledgements}
This work is supported by Hong Kong Research Grants Council General Research Fund (Project No. CUHK 412412 and Project No. CUHK 412513), a grant from the National Natural Science Foundation of China (Project No. 61233012) and a grant from Ministry of Science and Technology of the People's Republic of China under the Singapore-China 9th Joint Research Programme (Project No. 2013DFG12900).
The authors also gratefully thank the challenge organizers for helping the evaluation.

{\small
\bibliographystyle{ieee}
\bibliography{reference}

\begin{thebibliography}{10}\itemsep=-1pt

\bibitem{standardrank}
Standard competition ranking.
\newblock \url{https://en.wikipedia.org/wiki/Ranking}.

\bibitem{altunbay2010color}
D.~Altunbay, C.~Cigir, C.~Sokmensuer, and C.~Gunduz-Demir.
\newblock Color graphs for automated cancer diagnosis and grading.
\newblock {\em Biomedical Engineering, IEEE Transactions on}, 57(3):665--674,
  2010.

\bibitem{bertasius2015high}
G.~Bertasius, J.~Shi, and L.~Torresani.
\newblock High-for-low and low-for-high: Efficient boundary detection from deep
  object features and its applications to high-level vision.
\newblock In {\em ICCV}, pages 504--512, 2015.

\bibitem{chen2015automatic}
H.~Chen, Q.~Dou, D.~Ni, J.-Z. Cheng, J.~Qin, S.~Li, and P.-A. Heng.
\newblock Automatic fetal ultrasound standard plane detection using knowledge
  transferred recurrent neural networks.
\newblock In {\em MICCAI}, pages 507--514. Springer, 2015.

\bibitem{hao2016deep}
H.~Chen, X.~Qi, J.-Z. Cheng, and P.-A. Heng.
\newblock Deep contextual networks for neuronal structure segmentation.
\newblock In {\em Thirtieth AAAI Conference on Artificial Intelligence}, 2016.

\bibitem{chen2015spine}
H.~Chen, C.~Shen, J.~Qin, D.~Ni, L.~Shi, J.~C. Cheng, and P.-A. Heng.
\newblock Automatic localization and identification of vertebrae in spine {CT}
  via a joint learning model with deep neural networks.
\newblock In {\em MICCAI}, pages 515--522. Springer, 2015.

\bibitem{chen2014semantic}
L.-C. Chen, G.~Papandreou, I.~Kokkinos, K.~Murphy, and A.~L. Yuille.
\newblock Semantic image segmentation with deep convolutional nets and fully
  connected {CRF}s.
\newblock In {\em ICLR}, 2015.

\bibitem{ciresan2012deep}
D.~Ciresan, A.~Giusti, L.~M. Gambardella, and J.~Schmidhuber.
\newblock Deep neural networks segment neuronal membranes in electron
  microscopy images.
\newblock In {\em NIPS}, pages 2843--2851, 2012.

\bibitem{dai2015boxsup}
J.~Dai, K.~He, and J.~Sun.
\newblock Boxsup: Exploiting bounding boxes to supervise convolutional networks
  for semantic segmentation.
\newblock In {\em ICCV}, pages 1635--1643, 2015.

\bibitem{dhungel2015deep}
N.~Dhungel, G.~Carneiro, and A.~P. Bradley.
\newblock Deep learning and structured prediction for the segmentation of mass
  in mammograms.
\newblock In {\em MICCAI}, pages 605--612. Springer, 2015.

\bibitem{diamond2004use}
J.~Diamond, N.~H. Anderson, P.~H. Bartels, R.~Montironi, and P.~W. Hamilton.
\newblock The use of morphological characteristics and texture analysis in the
  identification of tissue composition in prostatic neoplasia.
\newblock {\em Human Pathology}, 35(9):1121--1131, 2004.

\bibitem{dou2016automatic}
Q.~Dou, H.~Chen, Y.~Lequan, L.~Zhao, J.~Qin, W.~Defeng, M.~Vincent, L.~Shi, and
  P.~A. Heng.
\newblock Automatic detection of cerebral microbleeds from {MR} images via {3D}
  convolutional neural networks.
\newblock {\em Medical Imaging, IEEE Transactions on}, 2016.

\bibitem{doyle2006boosting}
S.~Doyle, A.~Madabhushi, M.~Feldman, and J.~Tomaszeweski.
\newblock A boosting cascade for automated detection of prostate cancer from
  digitized histology.
\newblock In {\em MICCAI}, pages 504--511. Springer, 2006.

\bibitem{elston1991pathological}
C.~W. Elston, I.~O. Ellis, et~al.
\newblock Pathological prognostic factors in breast cancer. i. the value of
  histological grade in breast cancer: experience from a large study with
  long-term follow-up.
\newblock {\em Histopathology}, 19(5):403--410, 1991.

\bibitem{everingham2010pascal}
M.~Everingham, L.~Van~Gool, C.~K. Williams, J.~Winn, and A.~Zisserman.
\newblock The pascal visual object classes (voc) challenge.
\newblock {\em IJCV}, 88(2):303--338, 2010.

\bibitem{fakhrzadeh2012analyzing}
A.~Fakhrzadeh, E.~Sporndly-Nees, L.~Holm, and C.~L.~L. Hendriks.
\newblock Analyzing tubular tissue in histopathological thin sections.
\newblock In {\em Digital Image Computing Techniques and Applications (DICTA),
  2012 International Conference on}, pages 1--6. IEEE, 2012.

\bibitem{fleming2012colorectal}
M.~Fleming, S.~Ravula, S.~F. Tatishchev, and H.~L. Wang.
\newblock Colorectal carcinoma: pathologic aspects.
\newblock {\em Journal of gastrointestinal oncology}, 3(3):153--173, 2012.

\bibitem{fu2014novel}
H.~Fu, G.~Qiu, J.~Shu, and M.~Ilyas.
\newblock A novel polar space random field model for the detection of glandular
  structures.
\newblock {\em Medical Imaging, IEEE Transactions on}, 33(3):764--776, 2014.

\bibitem{gleason1992histologic}
D.~F. Gleason.
\newblock Histologic grading of prostate cancer: a perspective.
\newblock {\em Human pathology}, 23(3):273--279, 1992.

\bibitem{gunduz2010automatic}
C.~Gunduz-Demir, M.~Kandemir, A.~B. Tosun, and C.~Sokmensuer.
\newblock Automatic segmentation of colon glands using object-graphs.
\newblock {\em Medical image analysis}, 14(1):1--12, 2010.

\bibitem{he2015deep}
K.~He, X.~Zhang, S.~Ren, and J.~Sun.
\newblock Deep residual learning for image recognition.
\newblock {\em arXiv preprint arXiv:1512.03385}, 2015.

\bibitem{hinton2012improving}
G.~E. Hinton, N.~Srivastava, A.~Krizhevsky, I.~Sutskever, and R.~R.
  Salakhutdinov.
\newblock Improving neural networks by preventing co-adaptation of feature
  detectors.
\newblock {\em arXiv preprint arXiv:1207.0580}, 2012.

\bibitem{jacobs2014gleason}
J.~G. Jacobs, E.~Panagiotaki, and D.~C. Alexander.
\newblock Gleason grading of prostate tumours with max-margin conditional
  random fields.
\newblock In {\em Machine Learning in Medical Imaging}, pages 85--92. Springer,
  2014.

\bibitem{jia2014caffe}
Y.~Jia, E.~Shelhamer, J.~Donahue, S.~Karayev, J.~Long, R.~Girshick,
  S.~Guadarrama, and T.~Darrell.
\newblock Caffe: Convolutional architecture for fast feature embedding.
\newblock {\em arXiv preprint arXiv:1408.5093}, 2014.

\bibitem{krizhevsky2012imagenet}
A.~Krizhevsky, I.~Sutskever, and G.~E. Hinton.
\newblock Imagenet classification with deep convolutional neural networks.
\newblock In {\em NIPS}, pages 1097--1105, 2012.

\bibitem{leedeeply}
C.~Lee, S.~Xie, P.~Gallagher, Z.~Zhang, and Z.~Tu.
\newblock Deeply-supervised nets.
\newblock In {\em AISTATS}, 2015.

\bibitem{long2015fully}
J.~Long, E.~Shelhamer, and T.~Darrell.
\newblock Fully convolutional networks for semantic segmentation.
\newblock In {\em CVPR}, pages 3431--3440, 2015.

\bibitem{nguyen2012structure}
K.~Nguyen, A.~Sarkar, and A.~K. Jain.
\newblock Structure and context in prostatic gland segmentation and
  classification.
\newblock In {\em MICCAI}, pages 115--123. Springer, 2012.

\bibitem{qi2015semantic}
X.~Qi, J.~Shi, S.~Liu, R.~Liao, and J.~Jia.
\newblock Semantic segmentation with object clique potential.
\newblock In {\em ICCV}, pages 2587--2595, 2015.

\bibitem{ronneberger2015u}
O.~Ronneberger, P.~Fischer, and T.~Brox.
\newblock U-net: Convolutional networks for biomedical image segmentation.
\newblock In {\em MICCAI}, pages 234--241. Springer, 2015.

\bibitem{roth2015deeporgan}
H.~R. Roth, L.~Lu, A.~Farag, H.-C. Shin, J.~Liu, E.~B. Turkbey, and R.~M.
  Summers.
\newblock Deeporgan: Multi-level deep convolutional networks for automated
  pancreas segmentation.
\newblock In {\em MICCAI}, pages 556--564. Springer, 2015.

\bibitem{sabata2010automated}
B.~Sabata, B.~Babenko, R.~Monroe, and C.~Srinivas.
\newblock Automated analysis of pin-4 stained prostate needle biopsies.
\newblock In {\em Prostate Cancer Imaging}, pages 89--100. Springer, 2010.

\bibitem{shin2016deep}
H.-C. Shin, H.~R. Roth, M.~Gao, L.~Lu, Z.~Xu, I.~Nogues, J.~Yao, D.~Mollura,
  and R.~M. Summers.
\newblock Deep convolutional neural networks for computer-aided detection:
  {CNN} architectures, dataset characteristics and transfer learning.
\newblock {\em Medical Imaging, IEEE Transactions on}, 2016.

\bibitem{sirinukunwattana2016gland}
K.~Sirinukunwattana, J.~P. Pluim, H.~Chen, X.~Qi, P.-A. Heng, Y.~B. Guo, L.~Y.
  Wang, B.~J. Matuszewski, E.~Bruni, U.~Sanchez, et~al.
\newblock Gland segmentation in colon histology images: {The GlaS Challenge
  Contest}.
\newblock {\em arXiv preprint arXiv:1603.00275}, 2016.

\bibitem{sirinukunwattana2015stochastic}
K.~Sirinukunwattana, D.~Snead, and N.~Rajpoot.
\newblock A stochastic polygons model for glandular structures in colon
  histology images.
\newblock {\em Medical Imaging, IEEE Transactions on}, 34(11):2366 -- 2378,
  2015.

\bibitem{sirinukunwattana2015novel}
K.~Sirinukunwattana, D.~R. Snead, and N.~M. Rajpoot.
\newblock A novel texture descriptor for detection of glandular structures in
  colon histology images.
\newblock In {\em SPIE Medical Imaging}, pages 94200S--94200S. International
  Society for Optics and Photonics, 2015.

\bibitem{tabesh2007multifeature}
A.~Tabesh, M.~Teverovskiy, H.-Y. Pang, V.~P. Kumar, D.~Verbel, A.~Kotsianti,
  and O.~Saidi.
\newblock Multifeature prostate cancer diagnosis and gleason grading of
  histological images.
\newblock {\em Medical Imaging, IEEE Transactions on}, 26(10):1366--1378, 2007.

\bibitem{wu2005segmentation}
H.-S. WU, R.~Xu, N.~Harpaz, D.~Burstein, and J.~Gil.
\newblock Segmentation of intestinal gland images with iterative region
  growing.
\newblock {\em Journal of Microscopy}, 220(3):190--204, 2005.

\bibitem{xie2015holistically}
S.~Xie and Z.~Tu.
\newblock Holistically-nested edge detection.
\newblock In {\em ICCV}, pages 1395--1403, 2015.

\bibitem{yosinski2014transferable}
J.~Yosinski, J.~Clune, Y.~Bengio, and H.~Lipson.
\newblock How transferable are features in deep neural networks?
\newblock In {\em NIPS}, pages 3320--3328, 2014.

\bibitem{zhang2014facial}
Z.~Zhang, P.~Luo, C.~C. Loy, and X.~Tang.
\newblock Facial landmark detection by deep multi-task learning.
\newblock In {\em ECCV}, pages 94--108. Springer, 2014.

\end{thebibliography}
}

\end{document}